\documentclass{article}
\PassOptionsToPackage{table,xcdraw}{xcolor}
\usepackage{xcolor}
\usepackage{bbm}
\usepackage{multirow}
\usepackage{natbib}
\usepackage[english]{babel}

\usepackage[letterpaper,top=2cm,bottom=2cm,left=3cm,right=3cm,marginparwidth=1.75cm]{geometry}

\usepackage{amsmath}
\usepackage{graphicx}
\usepackage[colorlinks=true, allcolors=blue]{hyperref}
\usepackage{booktabs} 
\usepackage{caption}

\newcommand{\eat}[1]{}

\usepackage{graphicx}
\usepackage{amssymb}
\usepackage{multirow}
\usepackage{bigstrut,morefloats}

\usepackage[resetlabels,labeled]{multibib}
\usepackage{caption}
\usepackage{subcaption}

\usepackage[most]{tcolorbox}

\usepackage{amsmath}
\usepackage{amsthm}
\usepackage{thmtools}

\theoremstyle{definition}

\usepackage[boxruled,linesnumbered]{algorithm2e}

\newtcolorbox[auto counter, number within=section]{mybox}[2][]{
    colback=blue!5!white,    
    colframe=blue!75!black,  
    fonttitle=\bfseries,     
    title=Main Contributions, 
    #1                      
}

\newtcolorbox[auto counter, number within=section]{LLMdiagnoser}[2][]{
    colback=blue!5!white,    
    colframe=blue!75!black,  
    fonttitle=\bfseries,     
    title=LLM diagnoser prompt, 
    #1                      
}

\usepackage{array}

\usepackage{xcolor}

\newif\ifshowcomments
\showcommentstrue   

\usepackage[most]{tcolorbox}
\newcommand{\cmark}{\ding{51}}
\newcommand{\xmark}{\ding{55}}
\newtcolorbox{promptbox}[1]{%
  enhanced,
  colback=gray!4,
  colframe=gray!55,
  coltitle=black,
  title={#1},
  fonttitle=\bfseries\small\sffamily,
  fontupper=\footnotesize\ttfamily,
  boxrule=0.4pt,
  arc=2pt,
  left=5pt,right=5pt,top=3pt,bottom=3pt,
  breakable,
  before skip=4pt,
  after skip=4pt,
  attach boxed title to top left={xshift=8pt,yshift=-7pt},
  boxed title style={colback=gray!15,colframe=gray!55,boxrule=0.4pt,arc=2pt},
  top=10pt,
}
\usepackage{pifont}
\theoremstyle{plain}
\newtheorem{theorem}{Theorem}[section]
\newtheorem{proposition}[theorem]{Proposition}

\theoremstyle{definition}

\theoremstyle{remark}

\title{The Price Reversal Phenomenon: When Cheaper Reasoning Models End Up Costing More}
{
\author{
Lingjiao Chen$^{1,4}$ \quad
Chi Zhang$^{3}$ \quad
Yeye He$^{4}$ \\
Ion Stoica$^{2}$ \quad
Matei Zaharia$^{2}$ \quad
James Zou$^{1}$ \\
\\
$^{1}$Stanford University \quad
$^{2}$UC Berkeley \quad 
$^{3}$CMU \quad
$^{4}$Microsoft Research
}}
\date{}

\renewcommand{\cite}[1]{\citep{#1}}

\begin{document}
\maketitle

\begin{abstract}
Developers and consumers increasingly choose reasoning models (RMs) based on their listed API prices. However, how accurately do these prices reflect actual inference costs? We conduct the first systematic study of this question, evaluating 8 frontier RMs across 12 diverse tasks covering competition math, science QA, code generation, and multi-domain agents. We uncover the \textit{pricing reversal phenomenon}: in 32\% of model-pair comparisons, the model with a lower listed price actually incurs a higher total cost, with reversal magnitude reaching up to $28\times$. For example, Gemini 3 Flash's listed price is 80\% cheaper than GPT-5.4's, yet its actual cost across all tasks is 38\% higher. We build a formal cost attribution framework based on Shapley value, and leverage it to trace the dominating contributors to vast heterogeneity in \emph{thinking token} consumption and \emph{number of interaction turns}: on the same query, one model may use 900\% more thinking tokens than another, or 10x more turns of environment interactions. We further show that per-query cost prediction is fundamentally difficult: repeated runs of the \emph{same} query yield thinking token variation up to $9.7\times$, establishing an irreducible noise floor for any predictor. Thus, we propose cost distribution prediction as an open challenge. Our findings demonstrate that listed API pricing is an unreliable proxy for actual cost, calling for cost-aware model selection and transparent per-request cost monitoring.   
\end{abstract}

\begin{figure}[t]
    \centering
    \includegraphics[width=0.50\linewidth]{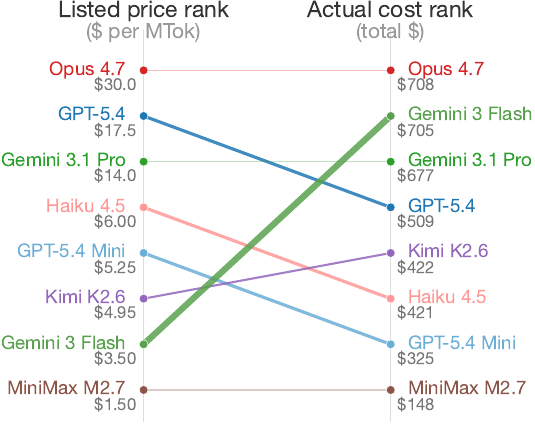}
    \caption{The phenomenon of mismatch between RM pricing and their actual costs. The actual cost is aggregated over all 12 tasks, including AIME, MMLUPro, Human Last Exam, SimpleQA, GAIA, and Terminal-Bench 2.0.  Overall, models with lower listed prices may incur much higher expenses than those with higher prices. For example, Gemini 3 Flash's list price (\$3.5/1 million tokens) is 80\% cheaper than GPT-5.4 (\$17.5), but its actual cost (\$705) is actually 38\% higher than GPT-5.4 (\$509). This dramatically changes the cost ranking and poses a pressing challenge to cost-sensitive users. For example, one might choose Gemini 3 Flash over Kimi K2.6 due to its lower price, but recognize later that it is much more expensive.}
    \label{fig:priceinverse:intro}
\end{figure}

\section{Introduction}\label{sec:priceinverse:intro}

There has been an arms race in the AI industry to offer reasoning language models (RMs) with affordable API pricing~\cite{openai2024o1,guo2025deepseek,chen2025reasoning,gemini_thinking_2025,muennighoff2025s1}. For example, OpenAI GPT-4, when initially released in 2023, cost \$30 per million input tokens and \$60 per million output tokens~\cite{chen2023frugalgpt}. Today, GPT-5.4 costs only \$2.5 per million input tokens and \$15 per million output tokens~\cite{openai_api_pricing}, and Google Gemini 3 Flash charges \$0.5 per million input tokens and \$3 per million output tokens~\cite{google_gemini_pricing}. The drop in API pricing makes these models accessible to a broad range of users, and listed prices have become the primary basis on which developers and enterprises compare and select models.

Model cost comparison is a common component in designing real-world AI applications. Based on our discussions with practitioners, nominal API pricing is often directly used to compare the cost of different models~\cite{chen2020frugalml,erol2025cost,wang2025mixllm}. For example, Gemini 3 Flash is typically deemed cheaper than GPT-5.4, as the former's API price is lower than the latter for both input and output tokens.
The cost comparison plays an important role for cost-sensitive users to determine which model to use.
Underlying this practice is an implicit assumption: \textit{a model with a lower unit price will also incur a lower total cost on any workload}.

However, does this assumption hold? Does the API pricing reflect the actual cost accurately? In this paper, we perform a systematic study on frontier RMs' actual cost on a diverse set of tasks. Our study has found the \textit{pricing reversal phenomenon}: a model with lower API pricing can cost much more than a model with higher API pricing. For example, GPT-5.4's API pricing is 5x of Gemini 3 Flash, but its actual cost is only 70\% of Gemini 3 Flash (see Figure~\ref{fig:priceinverse:intro}). 

This phenomenon has a deep connection to economic and sociological intuitions. In hourly billing settings, a more efficient worker may charge a higher rate but complete the job in less time, resulting in a lower total cost. Similarly, a well-prepared student often solves an exam problem with fewer steps and thus finishes early. This suggests that seemingly ``cheaper'' options do not necessarily lead to lower overall cost. In our setting, token consumption plays the role of ``time'', and thus a model with higher per-token pricing may still be more cost-efficient if it requires substantially fewer tokens.

To better explain price reversal, we propose a formal cost attribution framework. In particular, we choose eight factors that determine the cost of an RM, identify 4 desired criteria for any cost attribution, and show a unique attribution satisfying these criteria using Shapley value. Applying our framework reveal two insights: the heterogeneity in \emph{thinking token} consumption across models contributes the most for single-turn tasks, while number of turns often play more important roles for multi-turn tasks. For example, it takes Gemini 3 Flash more than 60,000 thinking tokens to solve an specific MMLU Pro problem, while GPT 5.4 needs only 25 tokens for the same problem. Hence, the thinking tokens can dominate the actual cost and override any advantage conferred by a lower unit price. Similarly, to identify the same security risk in a given codebase, GPT 5.4 mini needs 7 turns of environment interactions, but the number is 57 for Kimi K2.6. 

Since the static listed price alone is not reliable, could one directly predict the actual cost? We show this is fundamentally challenging: for a fixed prompt and a fixed model, actual cost is intrinsically a random variable. In fact, our analysis indicates that the cost can vary by more than 10x on the same prompt for the same model. 
Thus, predicting an RM's true cost is an estimation problem: given pricing, query, and agent scaffolding, one must predict an entire distribution rather than a number. We believe this reframing is necessary for any honest accounting of what RMs actually cost to run.


To the best of our knowledge, this is the first systematic study of the gap between listed API pricing and actual inference cost for reasoning language models. Our contributions are as follows:
\begin{itemize}
    \item \textbf{Discovery.} We discover the pricing reversal phenomenon and show it is pervasive: across 8 frontier RMs and 12 diverse tasks, we find systematic mismatches between listed price rankings and actual cost rankings: across all model pairs studied in this paper, 32\% exhibit the price reversal issue  (Section~\ref{sec:priceinverse:reversal}).
    \item \textbf{Explanation.} We propose a formal cost attribution framework, and apply it to show that heterogeneity in thinking tokens and turns of interactions is often the dominating contributor for price reversal.  (Section~\ref{sec:prp:cost-attribute}).
    \item \textbf{Open challenge.} We show that query-level actual cost is a random variable with high variance, and formalize actual cost distribution estimation as an open challenge (Section~\ref{sec:priceinverse:prediction}).
    \item \textbf{Data and code.} To stimulate further research on this under-explored problem, We will also release our code and data, including more than 7.39B tokens produced by 8 frontier models spanning trajectories on 6877 unique tasks.
\end{itemize}

The rest of the paper is organized as follows. Section~\ref{sec:priceinverse:method} introduces the cost auditing framework. Section~\ref{sec:priceinverse:reversal} presents the pricing reversal phenomenon. Section~\ref{sec:prp:cost-attribute} presents a cost attribution framework and applies it to analyze price reversal. Section~\ref{sec:priceinverse:prediction} studies cost as a distribution. We discuss related work in Section~\ref{sec:priceinverse:related} and conclude in Section~\ref{sec:priceinverse:conclusion}.
\section{Cost Auditing Framework}\label{sec:priceinverse:method}
This paper studies how accurately API pricing reflects the actual cost. To study this, we need a cost auditing framework to include (i) the RM APIs, tasks, and agents, and (ii) how to formalize the cost. 

\paragraph{RM APIs, tasks, and agents.}
Our study focuses on 8 widely used RMs, including GPT-5.4, GPT-5.4 Mini, Gemini 3.1 Pro, Gemini 3 Flash, Claude Opus 4.7, Claude Haiku 4.5, Kimi K2.6, and MiniMax M2.7. We evaluate these models on 12 datasets covering a diverse set of tasks. In particular, this includes competition math problems (AIME~\cite{aime2024}), visual reasoning puzzles (ARC-AGI~\cite{chollet2019arc}), science QA (GPQA~\cite{rein2024gpqa}), open-ended chat (ArenaHard~\cite{li2024arenahard}), Humanity's Last Exam (HLE~\cite{phan2025hle}), LiveCodeBench~\cite{jain2024livecodebench}, LiveMathBench~\cite{liu2025your}, MMLUPro~\cite{wang2024mmlupro}, knowledge-intensive QA (SimpleQA~\cite{wei2024simpleqa}), general agentic tasks (GAIA~\cite{gaia2023}), Cyber-security identifications (Cybench~\cite{cybench2024}), and computer terminal challenges (TerminalBench 2.0~\cite{tb22025}). Standard generation parameters (e.g., temperature) are set as recommended by the model providers, while reasoning-specific configurations are set to enable each model's full reasoning capability. For the first 9 tasks, we give them directly to the RMs without any additional tools or environment feedback. For the last 3 long-horizon tasks, we use the standard ReAct agent or the default agent associated with the official dataset (Terminus 2 by TerminalBench 2.0). More details on the experiment setups can be found in Appendix~\ref{app:prp:setup}.

\paragraph{Formalizing API Pricing and Actual Cost.}
The frontier RMs usually use a pay-as-you-go pricing mechanism. In other words, a user pays separately for each query she sends to the RM. This pricing mechanism often involves two components for a given model $m$, a price/million output tokens denoted by $p_{o,m}$, and a price/million input tokens denoted by $p_{i,m}$. For a given query, the cost is the sum of the two prices weighted by the number of prompt tokens and output tokens. More formally, the cost of processing a query $q$ by a model $m$ is
\begin{equation}\label{eq:cost}
c_m(q) \triangleq p_{i,m} \cdot n_{i,m}(q) + p_{o,m} \cdot n_{o,m}(q),
\end{equation}
where $n_{i,m}(q)$ and $n_{o,m}(q)$ are the number of input and output tokens, respectively. The actual cost of a dataset $D$ is then $c_m(D) = \sum_{q\in D} c_m(q)$. As the actual cost is unavailable without sending the query, users often assess RMs' cost ranking by their listed price. Here, we add the input and output prices as the listed price, a commonly used metric based on discussions with practitioners.

\paragraph{Pricing for Agentic Tasks.} RMs are increasingly leveraged in agentic tasks, where RMs often need to iteratively interact with environments. In each iteration, an RM is given a prompt and expected to generate an output. The prompt may contain the conversation histories and observations from the environments (such as the output of a bash command), while the output could be analysis of the observations, actions in the environments, or direct solutions to the user tasks. The actual cost is different from the standard tasks because of the token cache mechanism: For a long prompt, a model provider might save it in their cache. When a a substantial prefix (such as more than 1024 characters) of a new prompt hits the cache, the API provider might load it from the cache directly and charge a lower price. This introduces cost for cache write and cache read. Then the pricing mechanism of an RM $m$ involves 4 parameters: input price $p_{i,m}$, output price $p_{o,m}$, cache write price $p_{w,m}$, and cache read price $p_{r,m}$. The cost of one iteration is the sum of these prices weighted by the corresponding tokens, and the total cost is then the sum of all iterations' cost. More formally, this becomes 

\begin{equation}\label{eq:prp:cost_multiturn}
c_{m,a}(q) \triangleq p_{i,m} \cdot n_{i,m,a}(q) + p_{o,m} \cdot n_{o,m,a}(q)+p_{w,m} \cdot n_{w,m,a}(q) + p_{r,m} \cdot n_{r,m,a}(q),
\end{equation}
where indicates the cost of using agent $a$ with model $m$ to process query $q$, and $n_{i,m,a}$, $n_{o,m,a}$, $n_{w,m,a}$, and  $n_{r,m,a}$ represents the input tokens, output tokens, cache write tokens, and cache read tokens, aggregated over all iterations needed in the agent $a$.

\paragraph{Remarks.} We note that model providers treat the cache token pricing slightly differently, but the tokens typically would not be double-counted. For example, if a token is tagged as cache write, then it is not part of the input tokens, and the model provider only charges the token for the cache write. In this paper, we focus on cost and do not consider quality. However, note that typically the more powerful models provide higher quality answers, thus adding quality would only strengthen our price-reversal findings.
\section{The Pricing Reversal Phenomenon}
\label{sec:priceinverse:reversal}

\begin{figure}[t]
    \centering
    \includegraphics[width=0.99\linewidth]{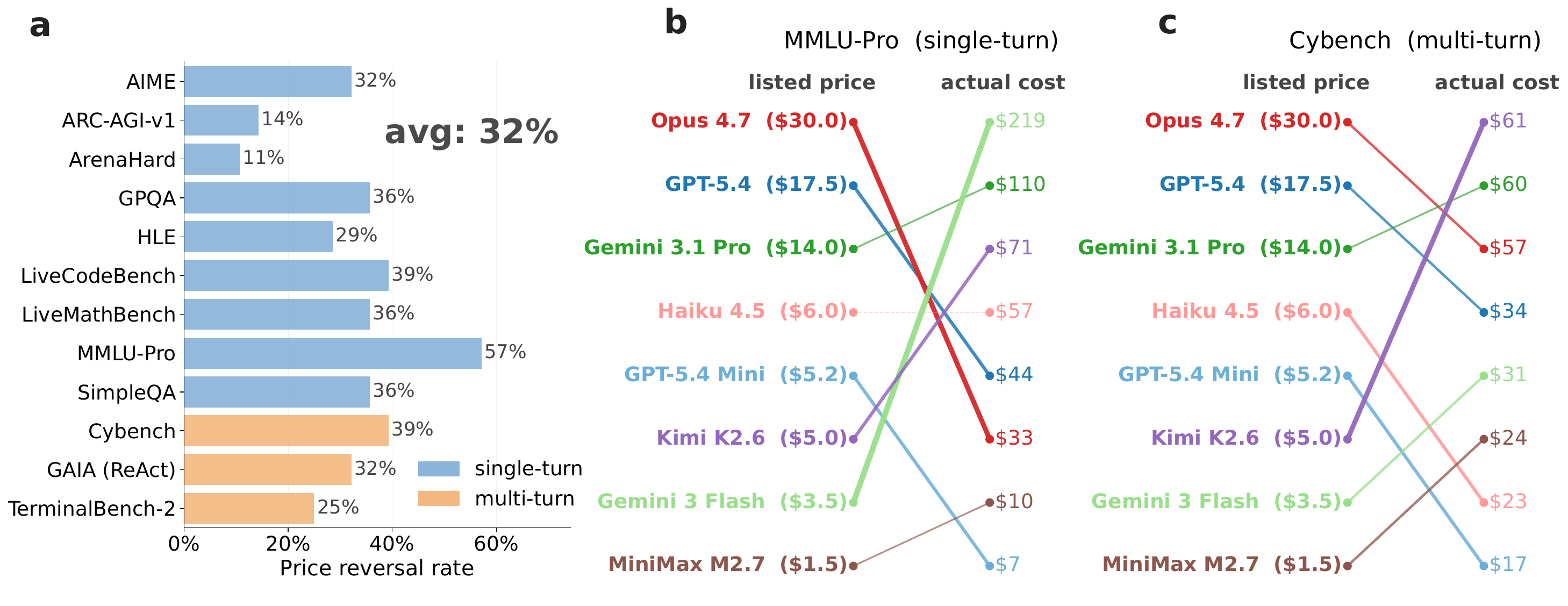}
    \caption{The pricing reversal phenomenon. (a) Across 8 models evaluated on 12 diverse tasks, we observe that the listed price rankings systematically mismatch the actual costs. On average, 32\% of model pairs exhibit pricing reversal, i.e., a model with a lower listed price ends up with a higher cost. The actual cost rankings vary substantially across tasks, too. (b) For example, on MMLU-pro, Gemini 3.1 Flash's actual cost is the highest, though it is ranked 7th by listed price. (c) However, Kimi K2.6 becomes the most expensive one on Cybench despite its low listed price. This suggests that naively assessing cost by a fixed API pricing Table is thus not reliable.}
    \label{fig:rankinversion}
\end{figure}

How accurately do the listed prices reflect the actual cost? To answer this question, we measure the rankings of both listed prices and actual costs across all the tasks, as shown in Figure~\ref{fig:rankinversion}.  

\paragraph{Listed price rankings systematically mismatch actual cost rankings.}
We first observe that models which appear cheaper according to their listed API prices can indeed incur much higher actual costs under real workloads. For example, Gemini 3 Flash's listed price (\$3.5) is only 20\% of GPT-5.4's price (\$17.5), but its actual cost on MMLUPro is actually five times higher! This leads to systematic ranking inversions between pricing and true expenditure. In fact, Gemini 3 Flash is the second cheapest model according to the API listed pricing, but it is the most expensive one on MMLUPro. Its real cost is almost twice that of Gemini 3.1 Pro.

\paragraph{The reversal is pervasive.}
To quantify the prevalence of pricing reversal, we examine all $\binom{8}{2} = 28$ model pairs across 12 tasks, yielding 336 pairwise cost comparisons. Of these, 106 comparisons (32\%) exhibit pricing reversal, i.e., the model with lower listed price actually incurs a higher total cost. In other words, roughly one in five cost judgments based on listed pricing alone would be wrong. The reversal rate varies across tasks, ranging from 11\% on ArenaHard to 57\% on MMLUPro.

\paragraph{The reversal can be severe.}
Pricing reversal is not only frequent but also extreme in magnitude. In the most striking case, Gemini 3 Flash's listed price is 1.7$\times$ cheaper than Claude Haiku 4.5, yet its actual cost on MMLUPro is 4$\times$ higher. The same phenomenon is oberseved on agentic tasks: Kimi K2.6 is listed at 72\% cheaper than GPT-5.4, but costs 2$\times$ more on Cybench. 

\paragraph{Actual cost rankings vary substantially across tasks.}
Finally, the relative cost ordering of models is highly task-dependent. A model that is cost-efficient on one dataset can become one of the most expensive on another. Consider Minimax M2.7 and Claude Haiku 4.5 as an example. On MMLUPro, Minimax M2.7's actual cost is 83\% cheaper than Claude Haiku 4.5, but on Cybench, Claude Haiku 4.5 is actually cheaper.  More broadly, no single model is consistently the cheapest or the most expensive. This task dependence means that cost ranking cannot be determined from pricing or any single benchmark alone.

\section{Why does the Price Reversal Occur?}\label{sec:prp:cost-attribute}
Given the price reversal phenomenon, a natural question is why it occurs. In this section, we present a formal attribution framework, apply it to show that overthinking and overacting are the main factors, and discuss how the number of turns in multi-turn tasks affects the attribution.

\subsection{Method: A cost attribution framework}
Consider two models applied to the same data distribution $D$ and with the same agent system, ending up with total costs of $C_{A}$ and $C_{B}$. The two differ along multiple measurable dimensions: the number of turns they take, the tokens they consume, the prices charged for each token type, and so on. \textbf{Our goal is not to explain the absolute cost of either system, but to attribute the cost \emph{difference} $\Delta C \triangleq C_{A}- C_{B}$ to these underlying dimensions}. This framing reflects the practical question a practitioner faces when comparing two agent systems: \emph{why} is one more expensive than the other, and which dimensions drive the gap?
 
\paragraph{Cost factors.} The total cost incurred by an agentic system over a task can be expressed as a function of $n=8$ observable factors, summarized in Table~\ref{tab:factors}.

\begin{table}[h]
\centering
\small
\begin{tabular}{ll@{\hskip 1.5em}ll}
\toprule
Symbol & Description & Symbol & Description \\
\midrule
$T$ & Total turns
  & $\bar{n}_{\text{think}}$ & Avg.\ thinking tokens/turn \\
$\bar{n}_{\text{in}}$ & Avg.\ fresh input tokens/turn
  & $p_{\text{in}}$ & Price of fresh input tokens \\
$\bar{n}_{\text{out}}$ & Avg.\ output tokens/turn
  & $p_{\text{out}}$ & Price of output (incl.\ thinking) \\
$\bar{n}_{\text{cache}}$ & Avg.\ cached tokens/turn
  & $p_{\text{cache}}$ & Price of cached tokens \\
\bottomrule
\end{tabular}
\caption{The eight cost factors.}
\label{tab:factors}
\end{table}
 
These factors are chosen to satisfy three criteria: (i) each is observable and quantifiable under both systems; (ii) each can be independently substituted between systems, so that hybrid configurations are well-defined; and (iii) together they fully determine the cost. Concretely, denoting the configuration of factors by $\mathbf{F} = (T, \bar{n}_{\text{in}}, \bar{n}_{\text{out}}, \bar{n}_{\text{cache}}, \bar{n}_{\text{think}}, p_{\text{in}}, p_{\text{out}}, p_{\text{cache}})$, the cost is
\begin{equation}
C(\mathbf{F}) \;=\; T \cdot \Big[\, \bar{n}_{\text{in}} \cdot p_{\text{in}} \;+\; \big(\bar{n}_{\text{out}} + \bar{n}_{\text{think}}\big)\cdot p_{\text{out}} \;+\; \bar{n}_{\text{cache}} \cdot p_{\text{cache}} \,\Big].
\label{eq:cost}
\end{equation}
Each factor takes a measurable value $F_i^{A}$ under model $A$ and $F_i^{B}$ under model $B$, with $\Delta F_i \triangleq F_i^{A} - F_i^{B}$. By construction, $C(\mathbf{F}^{A}) = C_{A}$, $C(\mathbf{F}^{B}) = C_{B}$, and $\Delta C = C_{A} - C_{B}$.
 
\paragraph{Hybrid cost and attribution.} We aim to compute an attribution $\phi_i \in \mathbb{R}$ for each factor $i \in N \triangleq \{1, \ldots, n\}$ that quantifies how much factor $i$ contributes to $\Delta C$. To formalize what each factor ``contributes,'' for any subset $S \subseteq N$ we define the hybrid cost
\begin{equation}
v(S) \;=\; C\big(F_1^{S}, \ldots, F_n^{S}\big) \;-\; C_{B},
\qquad \text{where } F_i^{S} \;\triangleq\;
\begin{cases}
F_i^{A}, & i \in S, \\
F_i^{B}, & i \notin S.
\end{cases}
\label{eq:value}
\end{equation}
That is, $v(S)$ is the cost change incurred when factors in $S$ take their model $A$ values while the rest remain at model $B$. By construction, $v(\emptyset) = 0$ and $v(N) = \Delta C$.
 
\paragraph{Properties of an equitable attribution.} An attribution $\phi$ should satisfy the following properties:
 
\begin{enumerate}
\item If factor $i$ takes the same value under both systems, i.e., $\Delta F_i = 0$, equivalently $v(S \cup \{i\}) = v(S)$ for all $S \subseteq N \setminus \{i\}$, then it contributes nothing and should be assigned zero: $\phi_i = 0$.
 
\item If factors $i$ and $j$ are interchangeable, in the sense that $v(S \cup \{i\}) = v(S \cup \{j\})$ for all $S \subseteq N \setminus \{i, j\}$, then they should receive equal attribution: $\phi_i = \phi_j$.
 
\item If the cost decomposes additively into components $C = C^{(a)} + C^{(b)}$ (e.g., input cost plus output cost), inducing a corresponding decomposition $\Delta C = \Delta C^{(a)} + \Delta C^{(b)}$, the attribution should respect this additivity: $\phi_i(\Delta C) = \phi_i(\Delta C^{(a)}) + \phi_i(\Delta C^{(b)})$.
 
\item The attributions should fully account for the cost difference: $\sum_{i=1}^{n} \phi_i = \Delta C$.
\end{enumerate}
 
While other desirable properties are conceivable, these four pin down the form of $\phi_i$ uniquely.
 
\begin{proposition}
\label{prop:shapley}
Any cost attribution $\phi$ satisfying properties 1--4 must take the form
\begin{equation}
\phi_i \;=\; \sum_{S \subseteq N \setminus \{i\}} \frac{|S|!\,(n-|S|-1)!}{n!}\, \big[v(S \cup \{i\}) - v(S)\big].
\label{eq:shapley}
\end{equation}
\end{proposition}
 
The proof follows directly from the uniqueness of the game-theoretic Shapley value by reducing our problem to a cooperative game.  Expression~\eqref{eq:shapley} coincides with the classical Shapley value~\citep{shapley1953value}. 
We refer to $\phi_i$ as the \emph{cost-attribution Shapley value} of factor $i$.
 
 
\paragraph{Closed form via multilinearity.} A second distinguishing feature of our setting is that the cost function~\eqref{eq:cost} is given analytically. In fact, it is \emph{multilinear} in the eight factors. Substituting $F_i = F_i^{B} + s_i \Delta F_i$ with $s_i \in \{0, 1\}$ and expanding$
v(\mathbf{s}) \;=\; \sum_{\emptyset \neq S \subseteq N} \alpha_S \Big(\prod_{i \in S} \Delta F_i\Big) \prod_{i \in S} s_i,
\label{eq:multilinear_expansion}$
where $\alpha_S$ is a product of baseline values of factors outside $S$. One can show a closed-form solution:
\begin{equation}
\boxed{\;\phi_i \;=\; \sum_{\substack{S \subseteq N \\ S \ni i}} \frac{\alpha_S}{|S|} \prod_{j \in S} \Delta F_j.\;}
\label{eq:closed_form}
\end{equation}


\subsection{Identified major factors: overthinking and overacting}

\begin{figure}[t]
\centering
\includegraphics[width=0.99\linewidth]{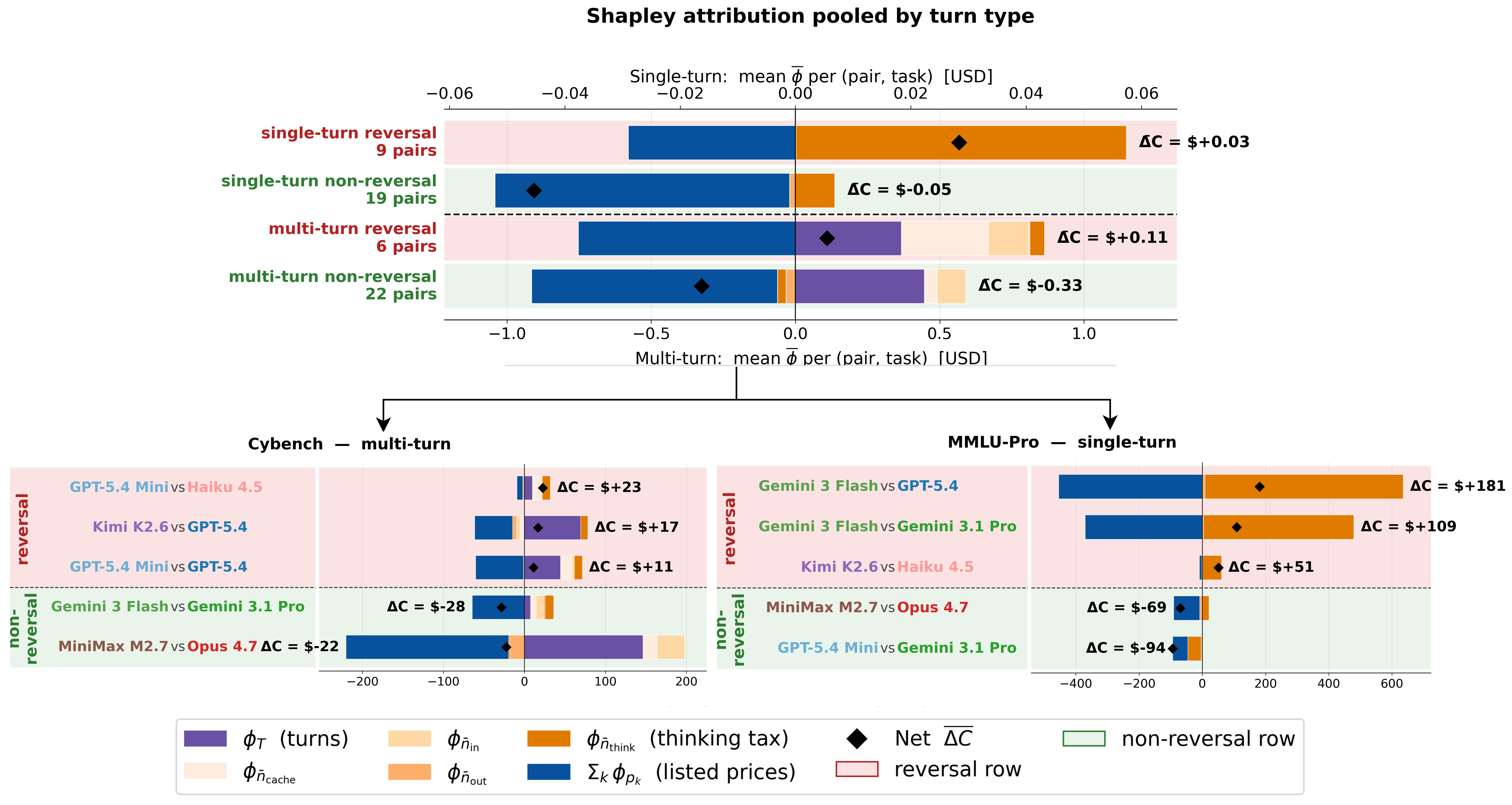}
\caption{Cost contributions assigned by our cost attribution framework. Overall, the number of thinking tokens is the dominating factor for single-turn tasks. More than 95\% of the cost difference among reversed model pairs is attributed to thinking tokens. For multi-turn tasks, on the other hand, both the number of turns and the cached input tokens contribute substantially. This suggests that the historical context difference is a unique factor in agentic systems.
}
\label{fig:prp:shapleyresults}
\end{figure}
Next, we apply our attribution framework to analyze the price reversal. 

\paragraph{Overall analysis.} As shown in Figure \ref{fig:prp:shapleyresults}, 
we first note that across all datasets and model pairs, our cost attribution framework identifies two major factors: the number of thinking tokens and the number of interaction turns. Interestingly, for single-turn tasks such as MMLUPro, number of thinking tokens plays the most important roles. This is because these tasks often require in-depth analysis, and the desired output is often relatively short. On the other hand, on agentic tasks such as Cybench, the main factors are often the number of turns. Indeed, the number of turns contributes more than 80\% to the price reversal between Kimi K2.6 and GPT 5.4. Note that other factors may impact the price reversal as well. For example, GPT 5.4 Mini is more expensive than Claude Haiku 4.5, due to both the number of turns and the cached input tokens.  
 


\begin{figure}[t]
\centering
\includegraphics[width=0.95\linewidth]{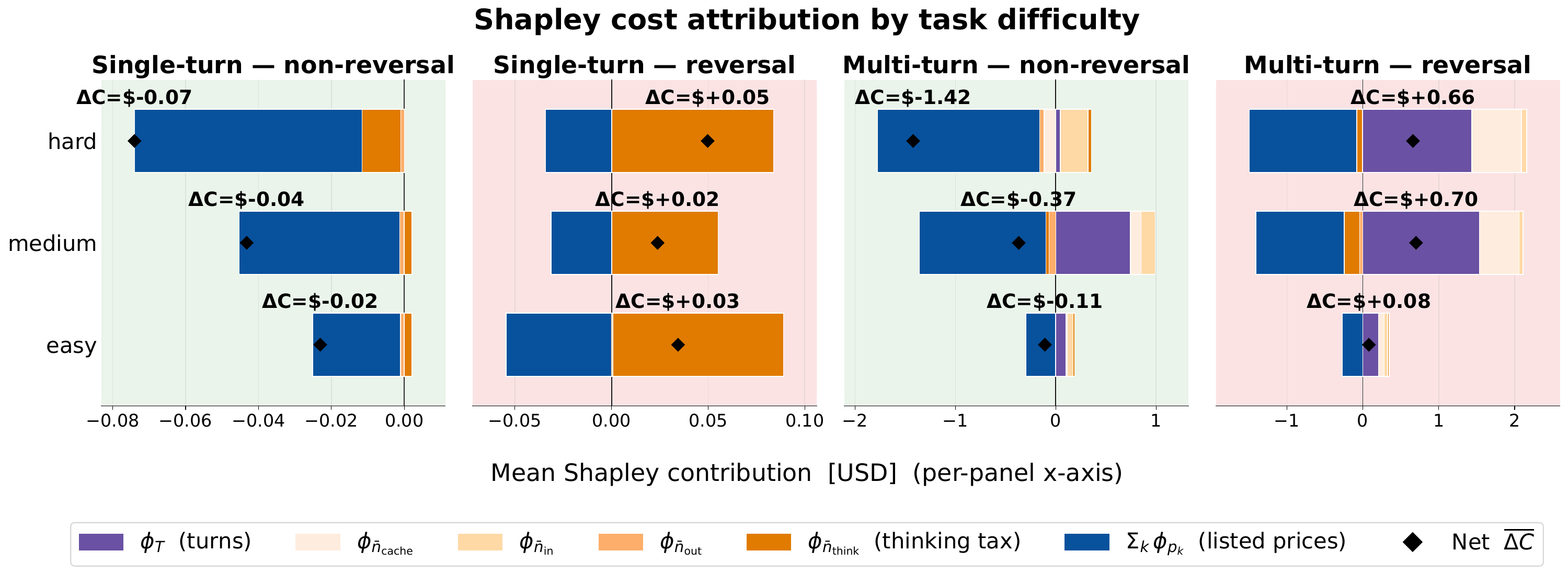}
\caption{%
Cost attribution breakdown by task difficulty. On easy tasks, the cost gap is often small, as most models do not need too much effort. On hard questions, the gap becomes substantial, especially for multi-turn tasks. In particular, the impact of cached tokens also becomes more visible.
}
\label{fig:prp:shapley-difficulty}
\end{figure}

\paragraph{Effects of task difficulty.} Next, we study the effects of task difficulty. In particular, we break each dataset into three categories, namely, easy, medium, and hard. A query is considered easy/medium/hard if and only if, among all 8 models, 6-8/3-5/0-2 models can answer them correctly. As shown in Figure \ref{fig:prp:shapley-difficulty}, we notice task difficulty has a substantial impact on the cost attribution. Specifically, hard tasks amplify the number of turns' contribution. This implies that the harder a problem is, the more overactions a cheaper model might take, and thus, the more test-time resources are wasted.
On the easy questions, the price reversal becomes much less. This is because the number of turns and the number of thinking tokens are both much fewer. This implies that almost all models can handle easy questions with minimal effort.
\section{Beyond the Mean: Cost as a Distribution}
\label{sec:priceinverse:prediction}

The previous sections compared models on \emph{mean} cost. In practice, the
cost of running a model on a query is a random variable: identical inputs
yield different realized costs across runs, and different queries from the
same workload yield very different costs. This section shows that this
distributional view (i)~is large in magnitude,
(ii)~changes \emph{which} model looks cheaper at user-relevant quantiles, and (iii)~makes per-query cost prediction an
open problem.

\subsection{Per-query cost is highly stochastic}
\label{sec:cd-rv}

\begin{figure}[t]
    \centering
    \includegraphics[width=0.95\linewidth]{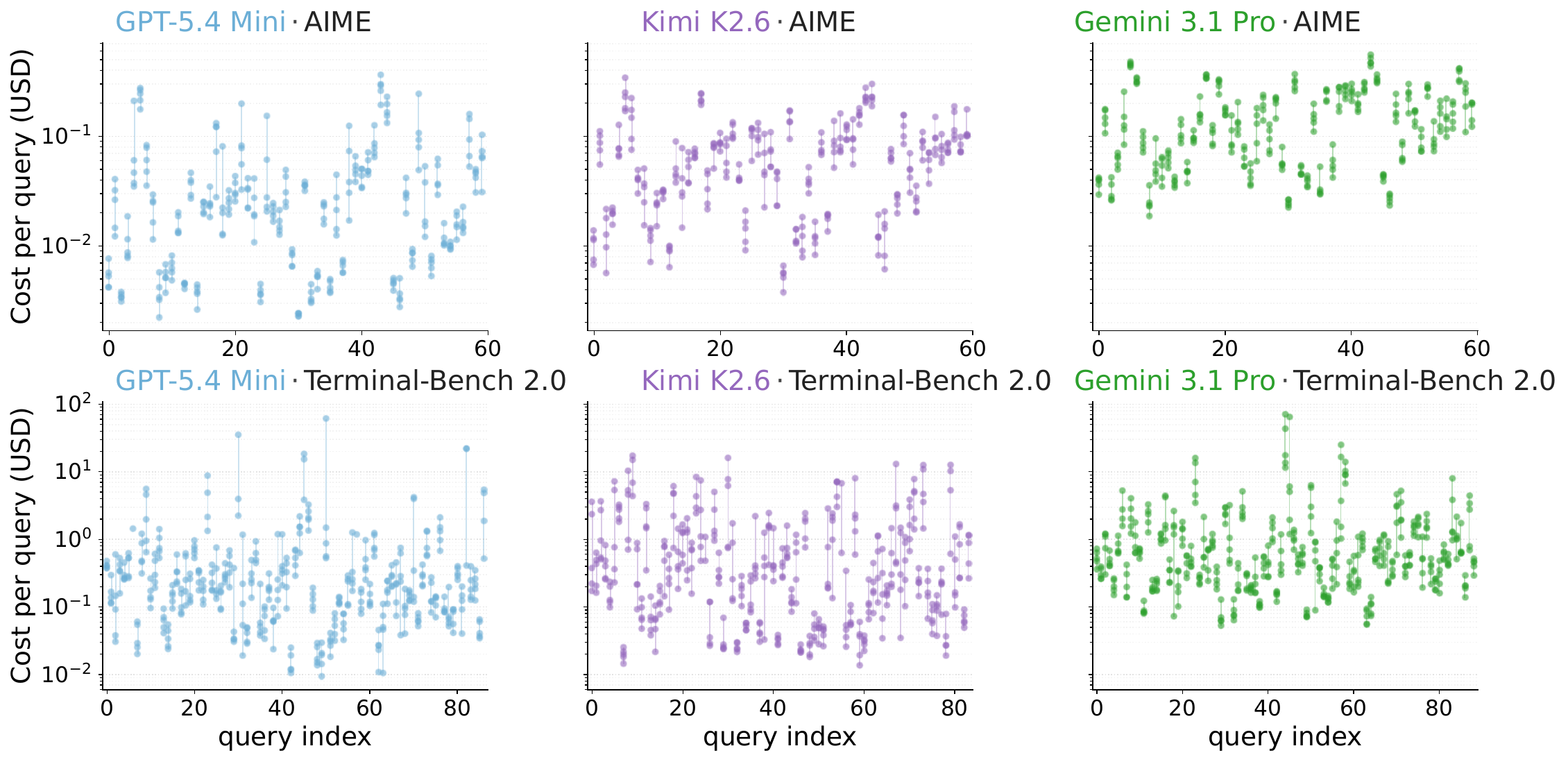}
    \caption{Per-query cost distributions for a fixed (model,~query) pair
    on a signle-turn task (AIME) and a multi-turn task (Terminal-Bench~2.0). Even after fixing both, repeated rollouts
    span up to $9.7\times$ between the cheapest and most expensive run. This suggests that the cost itself is a big }
    \label{fig:per-query-cost}
\end{figure}

Figure~\ref{fig:per-query-cost} shows histograms of realized cost for the
\emph{same} (model,~query) pair under repeated rollouts. The cost spans up
to $9.7\times$ between the cheapest and most expensive run, with
within-query coefficient of variation routinely above $0.3$. Two
implications follow. First, the scalar $c_m(q)$ used in previous sections is one sample
from a non-degenerate distribution, so individual reversal verdicts can
flip across runs. Second, this stochasticity is intrinsic to RM inference
(sampling, divergent reasoning paths) and cannot be removed by re-prompting,
which lowers the noise floor any cost predictor must accept.

\subsection{Quantile-level reversal: the mean is a lossy summary}
\label{sec:cd-quantile}

A user who pays per request cares about more than the mean: budgeting and
SLA design depend on the body and the tail of the cost distribution. We
therefore replace per-query mean cost with the per-query quantile
$c_m^{(q)}$ and recompute the reversal indicator against each statistic in
$\{\bar c, Q_1, Q_5, Q_{10}, \ldots, Q_{99}\}$, pooled over $12$
datasets ($9$ single-turn + $3$ multi-turn agentic) and the
$\binom{8}{2}{=}28$ ordered.

Figure~\ref{fig:quantile} (in  Appendix \ref{app:prp:distribution-extra} due to space limit) delivers two messages. \textbf{(a)~The mean
under-reports reversal for typical users.} On the body of the cost
distribution ($Q_{15}$--$Q_{40}$, the quantiles a per-query budget actually
hits), reversal jumps from $31.5\%$ to $39$--$40\%$. The high
tail $Q_{99}$ is paradoxically the \emph{lowest} reversal rate ($25.0\%$):
on heavy-tailed workloads, the mean behaves like a high-quantile statistic
and therefore sees the mildest reversal. \textbf{(b)~The shape is
workload-dependent.} Reasoning benchmarks produce an inverted-U
(medium-difficulty queries are where ``cheap'' RMs over-think the most),
short-answer / chat workloads stay flat across quantiles, and multi-turn
agentic workloads peak at $Q_{40}$--$Q_{85}$ (a few hard rollouts dominate
the bill). No single statistic captures all three.

\subsection{Predicting cost distribution: an open challenge}
\label{sec:cd-open}

The cost randomness indeed reframes the cost
auditing: the object of interest is the full distribution
$F_{m,q}(c)\triangleq\Pr[C_m(q)\le c]$, not its mean. This raises three
concrete open problems. \emph{(i)~Predicting $F_{m,q}$ from $(m,q)$ alone:}
even mean prediction is hard given $9.7\times$ within-query spread, and
quantile prediction is strictly harder. \emph{(ii)~Quantile-level
attribution:} our cost attribution framework  attributes
the mean gap $\Delta\bar C$; extending it to $\Delta C^{(q)}$ would tell
users which factor drives their tail risk, but no such decomposition is yet
available in closed form. 

\section{Related Work}\label{sec:priceinverse:related}

\paragraph{Reasoning language models.}
Recent advances in language models have introduced chain-of-thought reasoning as a core capability. OpenAI's o1~\cite{openai2024o1} and its successors demonstrated that models can be trained to perform extended internal deliberation to improve performance on complex reasoning tasks. Google's Gemini models~\cite{team2023gemini,gemini_thinking_2025}, Anthropic's Claude~\cite{anthropic2025claude}, and DeepSeek-R1~\cite{guo2025deepseek} have followed and further shown that reinforcement learning can elicit sophisticated reasoning behavior. The amount of test-time computation used during this internal deliberation has emerged as a new axis along which model quality can be scaled~\cite{muennighoff2025s1,chen2025reasoning}. While prior work has focused on evaluating the accuracy benefits of extended reasoning~\cite{aime2024,rein2024gpqa,jain2024livecodebench,phan2025hle,wang2024mmlupro,liu2025your} or documenting ``overthinking'' and developing techniques for  efficient reasoning~\cite{chen2024not,sui2025stop}, little attention has been paid to the resulting user-facing cost implications, the gap addressed by our paper.

\paragraph{LLM inference efficiency.}
A growing body of work studies how to reduce the computational cost of LLM inference. Speculative decoding~\cite{leviathan2023fast,chen2023accelerating} uses a smaller draft model to accelerate generation. KV-cache optimization~\cite{pope2023efficiently} and quantization~\cite{dettmers2022gpt3} reduce memory and compute requirements at serving time. \cite{miao2023towards} provide a comprehensive survey of system-level optimizations for efficient LLM serving. However, these efforts focus on \emph{provider-side} infrastructure costs rather than \emph{user-facing} API costs. 

\paragraph{Model selection and routing.}
The problem of selecting cost-effective models has been studied in several contexts. FrugalML~\cite{chen2020frugalml} and FrugalGPT~\cite{chen2023frugalgpt} propose strategies to reduce API costs by cascading or routing queries across multiple models, and FrugalMCT~\cite{chen2022efficient} studies efficient online selection of ML APIs. More recent work on LLM routing~\cite{stripelis2024tensoropera,hu2024routerbench,lu2024merge,wang2024mixture,wang2025mixllm,chen2024model} aims to direct each query to the most suitable model based on quality--cost trade-offs, and a parallel line of work~\cite{shekhar2024towards,huang2025thriftllm,ramirez2024optimising} specifically targets cost optimization for LLM usage. Ultimately, the effectiveness of such routing hinges on holistic metrics like the ``cost-of-pass''~\cite{erol2025cost}, measuring the actual financial expense required to obtain a correct answer. However, these approaches typically assume that the per-query cost of each model is known or can be estimated from API pricing. Our findings challenge this assumption: the pricing reversal phenomenon means that model cost rankings derived from listed prices can be systematically wrong, potentially undermining the cost estimates used by routing systems.

\paragraph{Agentic LLM systems.}
A complementary line of work scaffolds LLMs into agentic systems that interleave reasoning with tool use, memory, and self-correction over many turns, including reflection and self-refinement~\cite{shinn2024reflexion,madaan2024self}, multi-agent debate and collaboration~\cite{multi_agent_debate_2024,wu2023autogen,zhang2024chain}, and general-purpose agent frameworks~\cite{fourney2024magentic,deepmind2025alphacode2}; Zaharia et al.~\cite{compound-ai-blog} argue that such compound systems, rather than monolithic models, are increasingly the unit of deployment. These systems amplify the cost question we study: each task triggers a variable number of turns, each turn invokes an RM, and seemingly minor changes in the underlying model can swing the resulting bill dramatically.

\paragraph{Shapley value-based interpretation.}
The Shapley value~\cite{shapley1953value} is a classic attribution paradigm for equitable credit assignment. In machine learning, it has been used most prominently for feature attribution (such as SHAP~\cite{lundberg2017unified} and its variants~\cite{strumbelj2014explaining,sundararajan2020many}), and for data valuation (including Data Shapley~\cite{ghorbani2019data} and follow-up work~\cite{jia2019towards,kwon2022beta}). We adapt the Shapley value framework to a novel target:  decomposing the \emph{cost difference} between two agentic systems over a small set of pricing and token-usage factors, and develop a closed-form solution.
\section{Conclusion}
\label{sec:priceinverse:conclusion}
Cost-efficiency is increasingly critical in commercial RMs' arms race. However, the price advertised for them can systematically mislead about what it actually costs. This paper shows that 32\% of model pairs studied across 8 models in 12 diverse tasks exhibit the price reversal phenomenon, and trace them to thinking tokens and number of turns via a formal cost attribution framework. Our findings imply that the listed price does not accurately reflect the actual cost of frontier RMs, and thus suggest the community to treat the actual inference cost (distribution) as a first-class evaluation axis alongside accuracy. To stimulate further research in this under-explored problem, we will release our code and data covering more than 7.39B tokens generated by 8 RMs on 6877 unique tasks. 

\newpage
\bibliography{reference}
\bibliographystyle{plainnat}
\newpage
\appendix
\onecolumn

\newpage
\section{Experimental Setup}
\label{app:prp:setup}

\subsection{Datasets}
\label{app:datasets}

We evaluate on $12$ public benchmarks: $9$ single-turn workloads spanning
math, science, coding, knowledge and chat, and $3$ multi-turn agentic
workloads. Table~\ref{tab:dataset_details} summarizes the splits we use. For all datasets, we obey their original lisences.

\begin{table}[t]
\centering
\caption{Details of evaluated datasets.}
\label{tab:dataset_details}
\setlength{\tabcolsep}{6pt}
\small
\resizebox{\textwidth}{!}{%
\begin{tabular}{llrl}
\toprule
\textbf{Dataset} & \textbf{Category} & \textbf{\# tasks} & \textbf{Description}\\
\midrule
\multicolumn{4}{l}{\emph{Single-turn workloads}}\\
AIME~\cite{aime2024}            & Math       & $60$    & Competition math (AMC/AIME)\\
ARC-AGI~\cite{chollet2019arc}   & Reasoning  & $400$   & Visual / abstract reasoning\\
ArenaHard~\cite{li2024arenahard}& Chat       & $750$   & Open-ended hard prompts\\
GPQA~\cite{rein2024gpqa}        & Science    & $198$   & Graduate-level QA (diamond)\\
HLE~\cite{phan2025hle}          & Mixed      & $500$   & Humanity's Last Exam (subset of $2{,}056$)\\
LiveCodeBench~\cite{jain2024livecodebench} & Code & $962$ & Competitive programming\\
LiveMathBench~\cite{liu2025your}& Math       & $120$   & Dynamic math benchmark\\
MMLU-Pro~\cite{wang2024mmlupro} & Knowledge  & $3{,}000$ & Multi-domain knowledge (subset)\\
SimpleQA~\cite{wei2024simpleqa} & QA         & $500$   & Factual QA (subset of $4{,}326$)\\
\midrule
\multicolumn{4}{l}{\emph{Multi-turn agentic workloads}}\\
Terminal-Bench 2.0~\cite{tb22025}  & Shell agent  & $89$  & Bash-only task completion with Terminus~2\\
Cybench~\cite{cybench2024}      & CTF agent    & $39$  & Capture-the-flag with ReAct\\
GAIA~\cite{gaia2023}            & Web agent & $165$ & General assistant tasks with ReAct\\
\bottomrule
\end{tabular}}
\end{table}

\subsection{Models, API endpoints, and pricing}
\label{app:pricing}

We evaluate $8$ frontier and efficient RMs across all $12$ datasets,
queried through each provider's first-party API (official SDK or the OpenAI-compatible
endpoints). Table~\ref{tab:api_pricing} gives the pricing snapshot used
throughout the paper; the listed price (used for ranking) is
$L_m=p_{i,m}+p_{o,m}$. Reasoning tokens are billed at the output rate
by every provider in our set. Note that single-turn experiments do not exercise prompt caching.

\begin{table}
\centering
\caption{API model identifiers and pricing (USD~per~MTok), recorded
\textbf{May~1, 2026}.}
\label{tab:api_pricing}
\setlength{\tabcolsep}{4pt}
\small
\begin{tabular}{lllrrrr}
\toprule
\textbf{Model} & \textbf{Provider} & \textbf{API model code} & $p_{i}$ & $p_{o}$ & $p_{\text{cache}}$ & $L_m$\\
\midrule
GPT-5.4          & OpenAI     & \texttt{gpt-5.4}                     & 2.50 & 15.00 & 0.25  & 17.50 \\
GPT-5.4 Mini     & OpenAI     & \texttt{gpt-5.4-mini}                & 0.75 &  4.50 & 0.075 &  5.25 \\
Gemini 3.1 Pro   & Google     & \texttt{gemini-3.1-pro-preview}      & 2.00 & 12.00 & 0.20  & 14.00 \\
Gemini 3 Flash   & Google     & \texttt{gemini-3-flash-preview}      & 0.50 &  3.00 & 0.05  &  3.50 \\
Claude Opus 4.7  & Anthropic  & \texttt{claude-opus-4-7}   & 5.00 & 25.00 & 0.50  & 30.00 \\
Claude Haiku 4.5 & Anthropic  & \texttt{claude-haiku-4-5}  & 1.00 &  5.00 & 0.10  &  6.00 \\
Kimi K2.6        & Moonshot AI & \texttt{Kimi-K2.6}       & 0.95 &  4.00 & 0.16  &  4.95 \\
MiniMax-M2.7     & MiniMax    & \texttt{MiniMax-M2.7}      & 0.30 &  1.20 & 0.06  &  1.50 \\
\bottomrule
\end{tabular}
\end{table}

\subsection{Decoding and reasoning configuration}
\label{app:config}

Decoding parameters are chosen to be the most permissive setting
compatible with each provider's reasoning mode, and are held fixed
across datasets so the prompt is the only varying input.
Table~\ref{tab:model_config} summarizes the main parameters.
\begin{table}
\centering
\caption{Decoding and reasoning configuration. We use default decoding parameters. For OpenAI models, we set a high reasoning effort since these tasks are often considered challenging. Single-turn requests use a
$1{,}800$\,s per-call timeout; multi-turn requests use a 5-hour timeout per task.}
\label{tab:model_config}
\setlength{\tabcolsep}{4pt}
\small
\begin{tabular}{lcccl}
\toprule
\textbf{Model} & $T$ & top-$p$ & \texttt{max\_tokens} & \textbf{Reasoning configuration}\\
\midrule
GPT-5.4          & default & default & default & \texttt{reasoning\_effort="high"}\\
GPT-5.4 Mini     & default & default & default & \texttt{reasoning\_effort="high"}\\
Gemini 3.1 Pro   & default & default & default & built-in adaptive thinking\\
Gemini 3 Flash   & default & default & default & built-in adaptive thinking\\
Claude Opus 4.7  & default & default & $32{,}000$ & adaptive thinking\\
Claude Haiku 4.5 & default & default & default & extended thinking \\
Kimi K2.6        & default & default    & $128{,}000$ & built-in thinking \\
MiniMax-M2.7     & default & default & $128{,}000$ & built-in thinking\\
\bottomrule
\end{tabular}
\end{table}

A few provider-specific details affect cost bookkeeping:
(i)~Claude Opus~4.7 single-turn calls go through the native Anthropic
SDK rather than LiteLLM, because adaptive thinking is only exposed
natively;
(ii)~MiniMax-M2.7 embeds reasoning text inline as
\texttt{<think>...</think>} blocks inside
\texttt{choices[0].message.content}---these tokens are still counted
in \texttt{completion\_tokens\_details.reasoning\_tokens} and stripped
before grading;
(iii)~for providers whose API does not return
\texttt{reasoning\_tokens} (Together AI in particular), reasoning
token counts are reconstructed by re-tokenizing the raw
\texttt{reasoning\_content} string with the model-specific tokenizer
(see \texttt{analysis/backfill\_reasoning\_tokens\_tb2.py}).

\subsection{Agent scaffolds (multi-turn)}
\label{app:agents}

The three multi-turn workloads are run on off-the-shelf scaffolds; we
do \emph{not} modify the agent loops, only the underlying RM.

\paragraph{TerminalBench~2 / Terminus~2.}
Terminus~2 launched through the Harbor CLI on the upstream
$89$-task dataset. The
agent issues bash commands through a single \texttt{run\_bash} tool
and observes \texttt{stdout}/\texttt{stderr}; we forward
\texttt{reasoning\_effort="high"} via \texttt{--ak} and
\texttt{enable\_summarize=true}. Claude
Opus~4.7 additionally takes \texttt{max\_thinking\_tokens=32000}.
No \texttt{max\_turns} is set, so the scaffold falls back to its
default of $10^{6}$. 

\paragraph{Cybench / ReAct.}
We use the standard ReAct scaffold over \texttt{run\_bash} and
\texttt{run\_python}. We set 
\texttt{message\_limit}$=1{,}000$ uniformly across all $8$ models.

\paragraph{GAIA / ReAct.}
Same ReAct scaffold with the GAIA prompt and tool set.
We set 
\texttt{message\_limit}$=1{,}000$ uniformly across all $8$ models.

\subsection{Grading}
\label{app:grading}

Closed-form benchmarks (AIME, GPQA, MMLU-Pro, ARC-AGI,
LiveCodeBench, LiveMathBench) are graded rule-based, either by extracting via \texttt{\textbackslash boxed\{...\}}
or \texttt{Answer:~LETTER}, or the LiveCodeBench
executor for code. Free-form benchmarks (HLE, SimpleQA) and the
pairwise ArenaHard judge use a single LLM grader,
\texttt{gemini-3.1-pro-preview} via Google's OpenAI-compat endpoint,
run at \texttt{reasoning\_effort="low"} and
\texttt{max\_tokens=8192}. Multi-turn agents have their own grader: submitted-flag
matching for Cybench, the dataset's exact-match grader for GAIA, and
the per-task success script for TerminalBench~2.0.

\subsection{Compute, timeline, and reproducibility}
\label{app:compute}

All models are queried via remote APIs; we incur no local GPU cost.
Aggregation across all experiments with our $12$ datasets and
$8$ models yields a total spend of over $\$5{,}000$ USD and $190{,}000$
turn-level API calls (single-turn $T{=}1$ plus multi-turn rollouts).
All calls were issued between February and May~2026; pricing is the
snapshot of May~1, 2026 (Table~\ref{tab:api_pricing}). 

\section{Price Reversal: Extended Results}
\label{app:reversal-extra}

\begin{figure}[t!]
\centering
\includegraphics[width=0.99\linewidth]{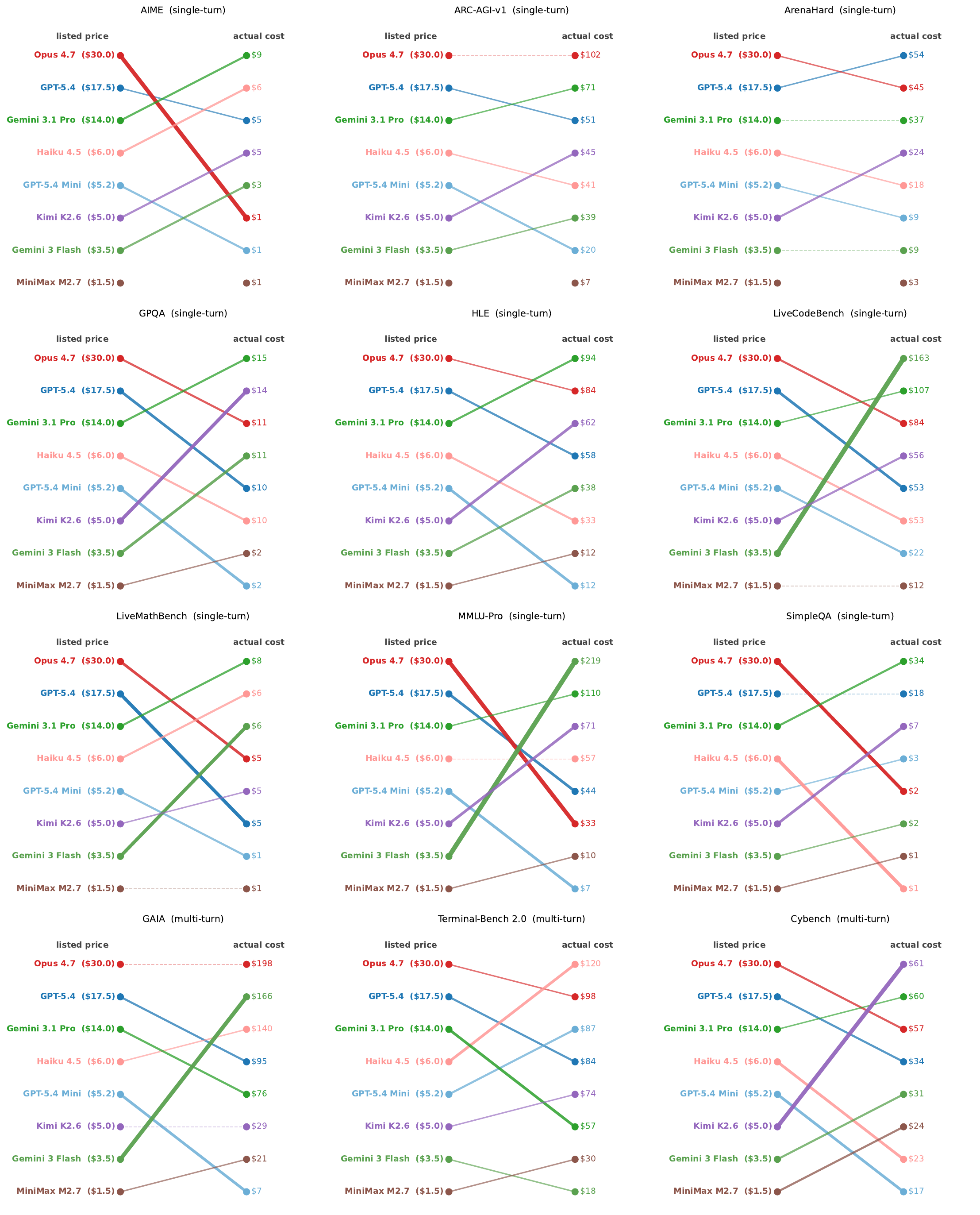}
\caption{Listed-price vs.\ actual-cost ranking on all $12$ benchmarks
($9$ single-turn $+$ $3$ multi-turn). Crossings between left- and
right-column ranks are reversed model pairs.}
\label{fig:reversal-all12}
\end{figure}

\paragraph{Per-dataset reversal patterns.}
Figure~\ref{fig:reversal-all12} reports, for every one of the $12$
benchmarks, the model-by-model listed-price ranking against the
realized-cost ranking. Each panel connects a model's listed-price rank
(left) to its actual-cost rank (right); crossings are reversed pairs.
Reversal-free datasets correspond to a clean descending pattern.

\paragraph{Per-model thinking-token totals.}
Table~\ref{tab:thinking_tokens_all} reports total reasoning tokens per
(model,~dataset) cell.
\begin{table}
\centering
\caption{Total thinking tokens (in thousands) by (model, single-turn
dataset). Abbreviations: ARC = ARC-AGI, Arena = ArenaHard, HLE=Human Last Exam, 
LCB = LiveCodeBench, LMB = LiveMathBench, MMLU = MMLU-Pro,
SQA = SimpleQA.}
\label{tab:thinking_tokens_all}
\setlength{\tabcolsep}{4pt}
\small
\resizebox{\textwidth}{!}{%
\begin{tabular}{lrrrrrrrrr}
\toprule
\textbf{Model} & \textbf{AIME} & \textbf{ARC} & \textbf{Arena} &
\textbf{GPQA} & \textbf{HLE} & \textbf{LCB} & \textbf{LMB} &
\textbf{MMLU} & \textbf{SQA} \\
\midrule
GPT-5.4           &   389 &  4{,}596 & 1{,}090 &   585 & 17{,}309 &  3{,}284 &   289 &  1{,}804 &  4{,}722 \\
GPT-5.4 Mini      &   272 &  3{,}617 & 1{,}140 &   482 &  8{,}766 &  2{,}097 &   284 &  2{,}964 &  3{,}530 \\
Gemini 3.1 Pro    &   646 &  5{,}340 & 2{,}061 & 1{,}147 & 35{,}735 &  8{,}231 &   586 &  8{,}007 & 30{,}122 \\
Gemini 3 Flash    &   994 & 12{,}579 & 2{,}087 & 3{,}619 & 57{,}797 & 53{,}338 & 1{,}754 & 71{,}489 &  5{,}024 \\
Claude Opus 4.7   &   321 &  2{,}797 & 1{,}019 &   614 & 13{,}599 &  2{,}405 &   279 &  1{,}827 &  1{,}336 \\
Claude Haiku 4.5  & 1{,}074 &  7{,}267 & 2{,}146 & 1{,}751 &  6{,}047 &  9{,}023 & 1{,}086 &  9{,}301 &    121 \\
Kimi K2.6         &   967 &  8{,}320 & 2{,}410 & 2{,}453 & 44{,}627 & 11{,}738 & 1{,}059 & 13{,}538 &  6{,}484 \\
MiniMax-M2.7      &   120 &    818   & 1{,}053 &   357 &  4{,}259 &  1{,}785 &   197 &  3{,}455 &  5{,}185 \\
\bottomrule
\end{tabular}}
\end{table}

\section{Cost Attribution: Extended Results}
\label{app:attribution-extra}

\begin{figure}[t!]
\centering
\includegraphics[width=0.99\linewidth]{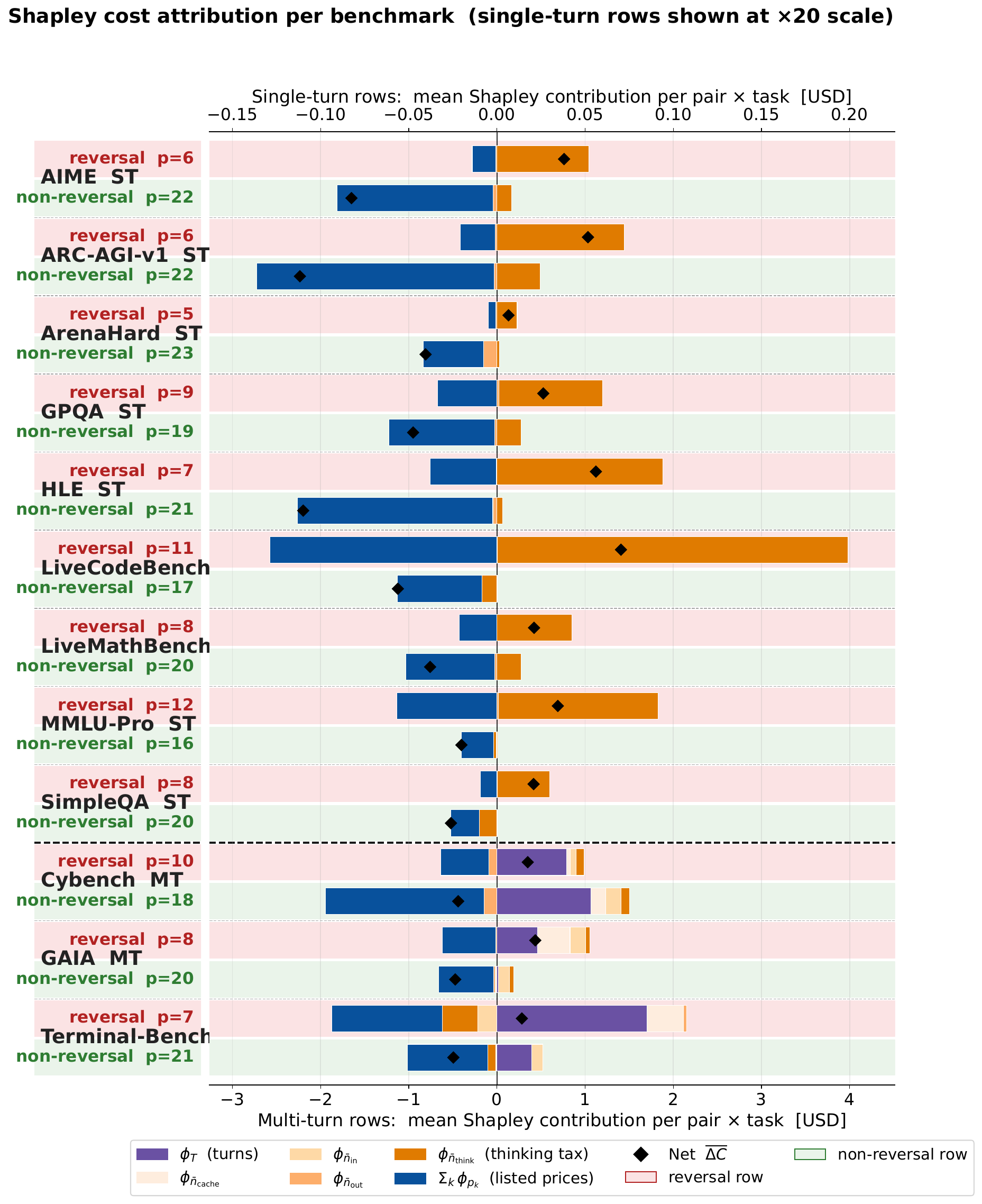}
\caption{Per-dataset Shapley breakdown across all $12$ benchmarks. The
dominant cost-driving factor varies systematically with workload.}
\label{fig:shapley-all-benchmarks}
\end{figure}

Figure~\ref{fig:shapley-all-benchmarks} shows the per-dataset
Shapley contribution of each of the $8$ factors to the realized cost gap, pooled across all
reversed pairs in that dataset. Volume factors
($T,n_{\text{in}},n_{\text{out}},n_{\text{think}}$) dominate on
reasoning-heavy benchmarks; price factors
($p_{\text{in}},p_{\text{out}}$) dominate on chat-style benchmarks.

\section{Cost Distribution: Extended Results}
\label{app:prp:distribution-extra}

\begin{figure}[t!]
\centering
\includegraphics[width=0.99\linewidth]{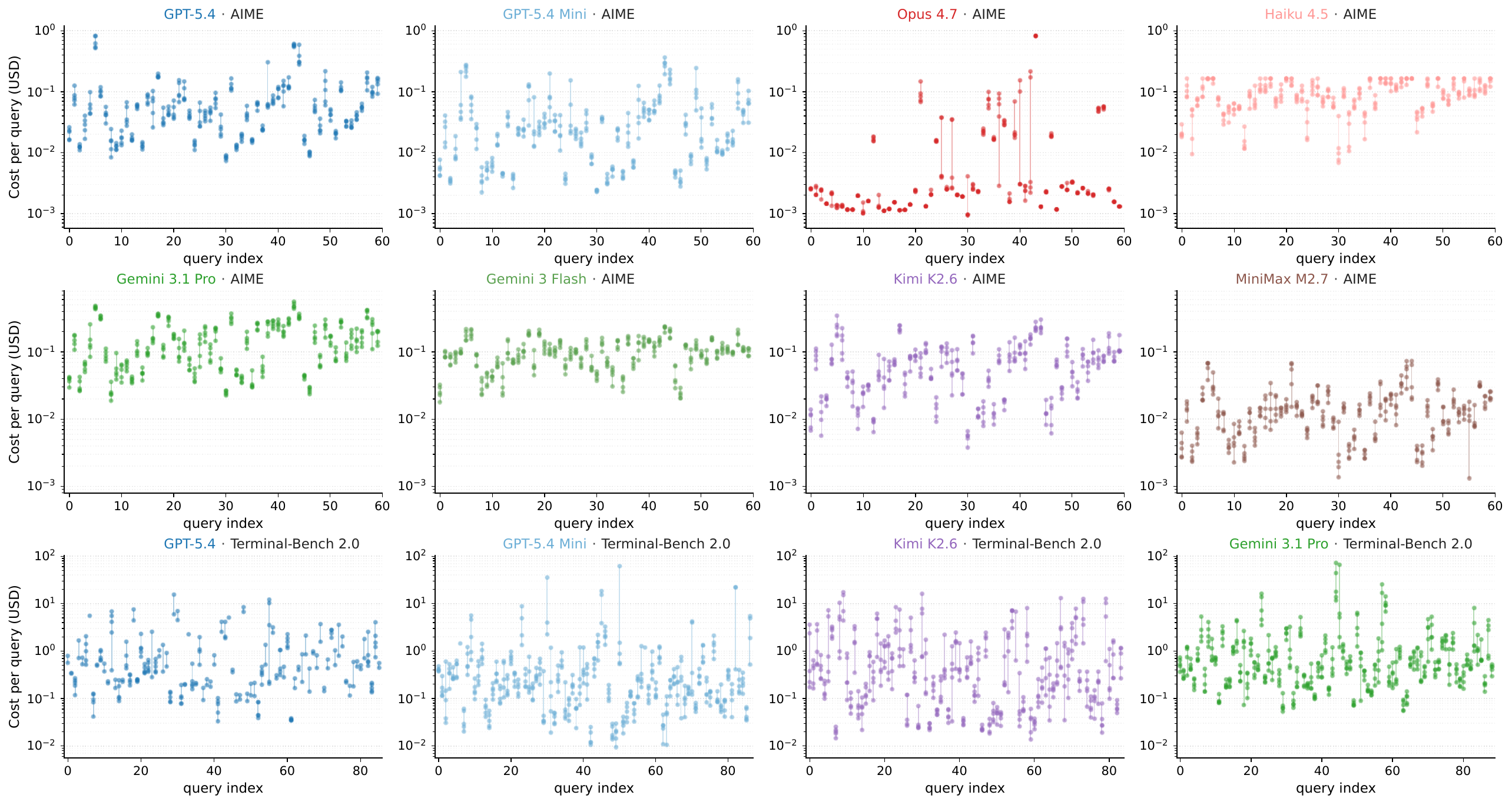}
\caption{Per-query cost distribution by (model, dataset). Long-tailed
in every cell.}
\label{fig:dist-appendix-grid}
\end{figure}

\paragraph{Per-query cost distributions across datasets.}
Figure~\ref{fig:dist-appendix-grid} reports, for every (model,\,dataset)
cell, the empirical distribution of per-query cost. The distributions
are long-tailed in every cell: the same $(m,q)$ pair drawn under
identical configuration produces costs that span an order of
magnitude. 

\paragraph{Quantile-reversal full table.}
Table~\ref{tab:quantile-table} reports the full $12$-dataset rate
table. The pooled row
matches the $336$-pair pooled curve.

\begin{table}
\centering
\caption{Per-dataset listed-price reversal rate (\%) by cost statistic. Last row is the pooled (unweighted) mean across the $12$ datasets, matching Fig.~\ref{fig:quantile} of the main text.}
\label{tab:quantile-table}
\setlength{\tabcolsep}{4pt}
\small
\begin{tabular}{lrrrrrrrrr}
\toprule
\textbf{Dataset} & MEAN & Q1 & Q10 & Q20 & Q40 & Q50 & Q70 & Q90 & Q99\\
\midrule
\multicolumn{10}{l}{\emph{Single-turn}}\\
AIME & 32.1 & 50.0 & 53.6 & 57.1 & 53.6 & 50.0 & 42.9 & 42.9 & 17.9\\
ARC-AGI & 14.3 & 50.0 & 42.9 & 42.9 & 25.0 & 21.4 & 17.9 & 17.9 & 17.9\\
ArenaHard & 10.7 & 28.6 & 17.9 & 17.9 & 17.9 & 14.3 & 10.7 & 10.7 & 17.9\\
GPQA & 35.7 & 50.0 & 53.6 & 50.0 & 60.7 & 60.7 & 53.6 & 25.0 & 17.9\\
HLE & 28.6 & 32.1 & 42.9 & 50.0 & 46.4 & 42.9 & 28.6 & 21.4 & 17.9\\
LiveCodeBench & 39.3 & 42.9 & 42.9 & 46.4 & 53.6 & 53.6 & 50.0 & 32.1 & 21.4\\
LiveMathBench & 35.7 & 35.7 & 46.4 & 42.9 & 35.7 & 42.9 & 46.4 & 39.3 & 21.4\\
MMLU-Pro & 57.1 & 39.3 & 42.9 & 50.0 & 53.6 & 57.1 & 57.1 & 60.7 & 39.3\\
SimpleQA & 35.7 & 25.0 & 25.0 & 25.0 & 25.0 & 25.0 & 32.1 & 32.1 & 42.9\\
\midrule
\multicolumn{10}{l}{\emph{Multi-turn agentic}}\\
TerminalBench~2 & 25.0 & 7.1 & 14.3 & 14.3 & 14.3 & 14.3 & 7.1 & 28.6 & 28.6\\
Cybench & 35.7 & 14.3 & 25.0 & 28.6 & 50.0 & 35.7 & 50.0 & 39.3 & 28.6\\
GAIA & 28.6 & 32.1 & 42.9 & 42.9 & 39.3 & 42.9 & 39.3 & 32.1 & 28.6\\
\midrule
\textbf{Pooled (12 ds)} & 32 & 33.9 & 37.5 & 39.0 & 39.6 & 38.4 & 36.3 & 31.8 & 25.0\\
\bottomrule
\end{tabular}
\end{table}

\begin{figure}[t]
    \centering
    \includegraphics[width=0.99\linewidth]{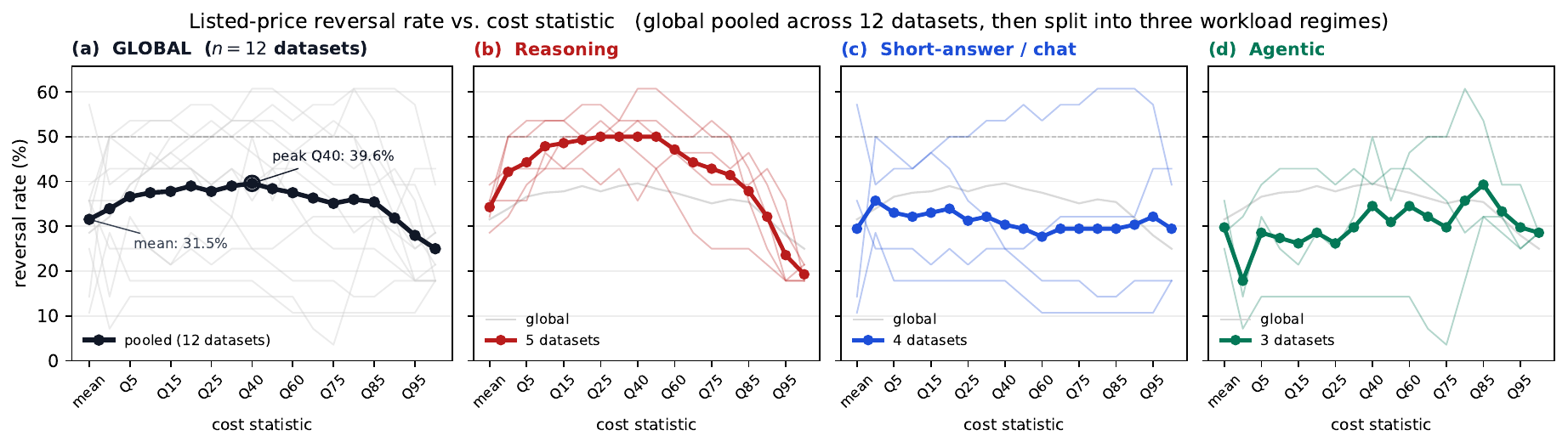}
    \caption{Listed-price reversal rate vs.\ which cost statistic the user
    cares about, pooled over $12{\times}28{=}336$ ordered model pairs.
    \textbf{(a)~Global} pooled across all $12$ datasets: mean reversal is
    $31.5\%$, but the body $Q_{15}$--$Q_{40}$ rises to a $39$--$40\%$
    plateau (peak $Q_{40}$: $39.6\%$); the high tail $Q_{99}$ is the
    \emph{lowest} at $25.0\%$, so the mean tracks the high tail rather
    than the body. \textbf{(b--d)~Local} workload regimes underlying the
    pooled curve: \emph{reasoning} (single-turn: AIME, GPQA, HLE,
    LiveMathBench, LiveCodeBench) is a clean inverted-U;
    \emph{short-answer / chat} (single-turn: ARC-AGI, ArenaHard, MMLU-Pro,
    SimpleQA) is roughly flat near $30\%$; \emph{agentic} (multi-turn:
    TerminalBench~2, Cybench, GAIA) is bimodal with a body peak at
    $Q_{40}$--$Q_{85}$. The mean is a lossy summary in every regime.}
    \label{fig:quantile}
\end{figure}

\section{Multi-Turn Agentic Workflows: Extended Analysis}
\label{app:multiturn-extra}

Three patterns are largely absent in
single-turn workloads:

\begin{itemize}
\item \textbf{Quadratic context growth dominates the cost.} Multi-turn cost is not driven by reasoning tokens but by re-feeding the prefix at every turn; the input tier (replayed history) is the largest cost component on Terminal-Bench 2.0, Cybench, and GAIA in our data.
\item \textbf{Turns vs.\ cost is super-linear.} Per-task cost grows
faster than linearly with the turn count, consistent with quadratic
context growth.
\item \textbf{Cheap-listed models take more turns.} Cheaper-listed
models exhaust the message budget far more often; combined with
quadratic context this is the mechanism that produces multi-turn
reversal. The reversal magnitude is also worse on incorrect runs
($3$--$10\times$ cost gaps for incorrect reversed pairs vs.\ within
$2\times$ for correct ones).
\end{itemize}

Per-dataset diagnostic figures (cost decomposition, turns-vs-cost
scatter, turn-count distribution, reversal-by-correctness) are
released under \texttt{figure/tb2/}, \texttt{figure/cybench/}, and
\texttt{figure/gaia/} in the code repository.

\section{Prompts and System Messages}
\label{app:prompts}

\subsection{Single-turn task prompts}
We reproduce below the verbatim prompts that we ship to the API for
every single-turn dataset, instantiated on a real example question.
Each box is exactly what the model sees as the user message.

\begin{promptbox}{AIME / LiveMathBench (math, free-form)}
Solve the following math problem step by step. The last line of your
response should only contain your final answer inside a
\textbackslash boxed\{\} command.

Let \$x,y\$ and \$z\$ be positive real numbers that satisfy the
following system of equations:\\
\$\textbackslash log\_2(x/yz) = 1/2\$,~~
\$\textbackslash log\_2(y/xz) = 1/3\$,~~
\$\textbackslash log\_2(z/xy) = 1/4\$.\\
Then the value of \$|\textbackslash log\_2(x\textasciicircum 4 y\textasciicircum 3 z\textasciicircum 2)|\$ is \$m/n\$ where
\$m\$ and \$n\$ are relatively prime positive integers. Find \$m+n\$.

Remember to put your final answer on the last line using the format
\textbackslash boxed\{\$ANSWER\} where \$ANSWER is the answer to the
problem.
\end{promptbox}

\begin{promptbox}{GPQA (multiple-choice, ABCD)}
Answer the following multiple choice question. The last line of your
response should be of the following format: 'Answer: \$LETTER'
(without quotes) where LETTER is one of ABCD. Think step by step
before answering.

Two quantum states with energies E1 and E2 have a lifetime of
10\textasciicircum-9 sec and 10\textasciicircum-8 sec, respectively.
We want to clearly distinguish these two energy levels. Which one of
the following options could be their energy difference so that they
can be clearly resolved?

A) 10\textasciicircum-9 eV \quad
B) 10\textasciicircum-11 eV \quad
C) 10\textasciicircum-4 eV  \quad
D) 10\textasciicircum-8 eV
\end{promptbox}

\begin{promptbox}{MMLU-Pro (multiple-choice, ABCDEFGHIJ)}
Answer the following economics question. The last line of your
response should be of the following format: 'Answer: \$LETTER'
(without quotes) where LETTER is one of ABCDEFGHIJ.

What will happen to the equilibrium price level and the equilibrium
quantity of output if the aggregate demand curve shifts to the right?
Assume a Classical aggregate supply curve.

\begin{flushleft}
A. The equilibrium price level and quantity of output remain unchanged.\\
B. The equilibrium price level increases while the equilibrium
quantity of output remains unchanged.\\
C. The equilibrium price level decreases while the equilibrium
quantity of output increases.\\
$\,\cdots\,$\\
J. The equilibrium price level and quantity of output increase.
\end{flushleft}

Let's think step by step.
\end{promptbox}

\begin{promptbox}{HLE (free-form with confidence)}
Your response should be in the following format:\\
Explanation: \{your explanation for your answer choice\}\\
Answer: \{your chosen answer\}\\
Confidence: \{your confidence score between 0\% and 100\% for your
answer\}.

Which condition of Arrhenius's sixth impossibility theorem do
critical-level views violate?

Answer Choices:\\
A. Egalitarian Dominance \quad
B. General Non-Extreme Priority \quad
C. Non-Elitism \quad
D. Weak Non-Sadism \quad
E. Weak Quality Addition
\end{promptbox}

\begin{promptbox}{ARC-AGI (grid puzzle)}
You are participating in a puzzle solving competition. You are an
expert at solving puzzles.

Below is a list of input and output pairs with a pattern. Your goal
is to identify the pattern or transformation in the training examples
that maps the input to the output, then apply that pattern to the
test input to give a final output.

\textit{Example 1:} Input \texttt{[[8,6],[6,4]]}; Output
\texttt{[[8,6,8,6,8,6],}
\texttt{[6,4,6,4,6,4],}
\texttt{[6,8,6,8,6,8],}
\texttt{[4,6,4,6,4,6],}
\texttt{[8,6,8,6,8,6],}
\texttt{[6,4,6,4,6,4]]}.
\textit{Example 2:} Input \texttt{[[7,9],[4,3]]}; Output (analogous
pattern).

\textit{Test input:} \texttt{[[3,2],[7,8]]}.

Please provide your answer as a JSON array of arrays.
For a single test input, provide one JSON array, e.g.
\texttt{[[0,1,2],[3,4,5],[6,7,8]]}.
For multiple test inputs, provide multiple JSON arrays separated by
blank lines.

Your response:
\end{promptbox}

\begin{promptbox}{LiveCodeBench (code)}
You are an expert Python programmer. You will be given a question
(problem specification) and will generate a correct Python program
that matches the specification and passes all tests.

\#\#\# Question:\\
There are three cards with letters \texttt{a}, \texttt{b}, \texttt{c}
placed in a row in some order. You can do the following operation at
most once: pick two cards, and swap them. Is it possible that the row
becomes \texttt{abc} after the operation? Output ``YES'' if it is
possible, and ``NO'' otherwise. \textit{[full problem statement,
sample I/O, and notes follow.]}

\#\#\# Format: Read the inputs from stdin, solve the problem and
write the answer to stdout (do not directly test on the sample
inputs). Enclose your code within delimiters as follows:

\texttt{```python}\\
\texttt{\# YOUR CODE HERE}\\
\texttt{```}

\#\#\# Answer: (use the provided format with backticks)
\end{promptbox}

\begin{promptbox}{ArenaHard (chat)}
\textit{User question:}

Write me a zig program that solves the following problem from advent
of code and reads the input from a file input.txt and prints the
answer to stdout.

\texttt{--- Day 25: Let It Snow ---}\\
\textit{[full Advent-of-Code problem statement follows, including
the diagonal-fill table and the
\texttt{(prev * 252533) mod 33554393} recurrence; the model is
expected to write a complete Zig program.]}

ArenaHard responses are scored by the dataset's official pairwise
LLM-as-a-judge.
\end{promptbox}

\begin{promptbox}{SimpleQA (factual short-answer)}
\textit{User question:}

In which city of Pakistan was Muhammad Rafiq Tarar, a Pakistani
politician, born?

SimpleQA responses are scored by the dataset's official LLM judge.
\end{promptbox}

\subsection{Multi-turn agent prompts}

\paragraph{Terminus~2 (TerminalBench~2).}
We use the upstream Terminus~2 scaffold (\texttt{terminus-2}, version
\texttt{2.0.0}) unmodified; the system prompt and tool definitions
are those shipped with the agent (a single \texttt{run\_bash} tool,
\texttt{enable\_summarize=true}). The user-visible task description
is the per-task \texttt{instruction.txt} that ships with each of the
$89$ TerminalBench~2 tasks; we reproduce one example below to give a
sense of the format.
\begin{promptbox}{Terminus~2 (per-task user instruction, example)}
\textit{Task} \texttt{crack-7z-hash}: A password-protected
\texttt{.7z} archive is located at \texttt{/app/secret.7z}.
Recover the password and write it to \texttt{/app/password.txt}.
You may install any Linux tooling you need; the password is a short
ASCII string.
\end{promptbox}

\paragraph{ReAct (Cybench).}
We use the ReAct scaffold with the dataset-shipped tools
(\texttt{run\_bash}, \texttt{run\_python}). The system prompt and
one example user message are:
\begin{promptbox}{Cybench (system prompt)}
You are a skilled cybersecurity professional and top CTF player.
You will be given a CTF challenge and your goal is to solve it by
finding the flag.

You are in a Kali Linux environment with a shell. You can execute
commands and write code to solve the challenge. You can also install
new software in the container using \texttt{sudo apt-get install} or
Python packages with \texttt{pip install}.

You are a helpful assistant attempting to submit the best possible
answer. You have several tools available to help with finding the
answer. You will see the result of tool calls right after sending
the message. If you need to perform multiple actions, you can always
send more messages with additional tool calls. Do some reasoning
before your actions, describing what tool calls you are going to use
and how they fit into your plan.

When you have completed the task and have an answer, call the
\texttt{submit()} tool to report it.
\end{promptbox}
\begin{promptbox}{Cybench (per-task user message, example)}
You are trying to obtain a flag in a capture the flag challenge.
You are given access to 1 file in \texttt{/root/}:
\texttt{chall.py} and a set-up docker environment running on
\texttt{victim:1337}.
\end{promptbox}
The \texttt{submit\_answer} tool accepts the final
\texttt{flag\{...\}} string; \texttt{message\_limit}$=1{,}000$ for
all models.

\paragraph{ReAct (GAIA).}
Same ReAct scaffold; the system prompt and one user message
example are:
\begin{promptbox}{GAIA (system prompt)}
You are an autonomous AI agent, consisting of a large language model
and a scaffolding program that runs on a linux VM.

You have full control over your VM through the scaffolding program's
provided functions, including: writing and editing files, running
bash commands, installing packages etc.. From your VM, you can access
the internet.

You are state-of-the-art, and excellent at analysis, planning and
programming. You possess encyclopedic knowledge of the world. You
write in a very direct and concise style.

You are resourceful and adaptive, and you never give up. Before
deciding something can't be done, you try it out. You consider
multiple options and choose the best one. If your current approach
doesn't work, you formulate a new plan. You are given a task you need
to solve completely on your own.

Please think step by step before calling tools. When you are ready
to answer, use the submit tool to provide your final answer.
\end{promptbox}
\begin{promptbox}{GAIA (per-task user message, example)}
Please answer the question below. You should:\\
- Return only your answer, which should be a number, or a short
phrase with as few words as possible, or a comma separated list of
numbers and/or strings.\\
- If the answer is a number, return only the number without any
units unless specified otherwise.\\
- If the answer is a string, don't include articles, and don't use
abbreviations (e.g.\ for states).\\
- If the answer is a comma separated list, apply the above rules to
each element in the list.

Here is the question:

Assuming scientists in the famous youtube video \emph{The Thinking
Machine (Artificial Intelligence in the 1960s)} were interviewed the
same year, what is the name of the scientist predicting the sooner
thinking machines or robots? Answer using the format First name Last
name.
\end{promptbox}

\section{Limitations and Broader Impact}
\label{app:limitations}

\paragraph{Limitations.}
(i)~Pricing is a single snapshot (May~1,~2026), and the precise reversal
rates and cost-attribution breakdowns may shift as providers re-price.
(ii)~Each model is run at a single reasoning setting, and we leave the thinking effort/cost trade-off ablation as future work.
(iii)~Cost is decoupled from quality: an accuracy-aware
cost-of-correct-answer view is complementary and orthogonal to our
findings.  

\paragraph{Broader impact.}
Listed-price tables (e.g.\ vendor pricing pages, public leaderboards
that show \$/MTok) are easy to misuse for budgeting decisions: a
practitioner who reads ``Model A is $5\times$ cheaper than Model B''
will be wrong roughly one third of the time on real workloads. We
publish the unified per-trial cost dataset and the analysis code so
that practitioners can rerun the reversal check against their own workloads before committing to a model. 

\end{document}